\documentclass{IEEEtran}

\usepackage[utf8]{inputenc}
\usepackage{amsmath,amssymb}
\usepackage{graphicx}
\usepackage{booktabs}
\usepackage{cite}
\usepackage[colorlinks=false]{hyperref}
\usepackage{multirow}

\title{\textbf{CALOS}: \textbf{C}ontrol-\textbf{A}ffine \textbf{L}yapunov \textbf{O}n-manifold \textbf{S}afety Layer for Safe Deep Reinforcement Learning for Quadrotors}

\author{Fabrizio Cesareo, Sebastiano Mengozzi, Nicola Mimmo, Andrea Acquaviva\\[4pt]
\normalsize Department of Electrical, Electronic, and Information Engineering ``Guglielmo Marconi'' --- DEI\\
\normalsize University of Bologna, Italy}

\begin{document}
\maketitle

\begin{abstract}
\textbf{Deep Reinforcement Learning has demonstrated remarkable capability in quadrotor control, yet learned policies offer no guarantee of respecting safety constraints during training or deployment.
We present CALOS (Control-Affine Lyapunov On-manifold Safety), a runtime safety layer that enforces attitude constraints on a quadrotor without modifying the underlying learning algorithm.
CALOS formulates four tilt-angle inequalities and a Lyapunov descent condition as a single quadratic program whose solution is the minimum-norm correction to the nominal torque output of the policy.
The quadratic program is solved exactly via active-set enumeration over the three-dimensional torque space, with a computational cost low enough to enforce constraints in real time across thousands of parallel simulation environments, as required by modern massively parallel Deep Reinforcement Learning training.
Evaluated on trajectory-tracking tasks in NVIDIA Isaac Lab, CALOS reduces lateral tracking error by 55--60\% relative to an unconstrained Proximal
Policy Optimization baseline while achieving zero attitude-constraint violations on the training trajectory. By restricting exploration to safe regions of the state space, the safety layer also accelerates training convergence and improves data efficiency without producing suboptimal policies.}
\end{abstract}

\section{Introduction}
\label{sec:intro}

Deep Reinforcement Learning (DRL) policies trained in simulation can push quadrotors to the edge of their flight envelope~\cite{Kaufmann2023}, yet the resulting controllers inherit no classical stability guarantee~\cite{khalil2002nonlinear}. Safe DRL~\cite{Garcia2015} addresses this gap through two broad strategies: modifying the optimization criterion via Constrained Markov Decision Processes (CMDPs)~\cite{Altman1999,chow2018lyapunov}, or modifying the action at runtime via shielding~\cite{alshiekh2018shielding} or control barrier functions~\cite{cheng2019end}. A persistent tension remains: CMDP solvers can destabilize learning~\cite{tessler2019reward}, while runtime filters applied during training risk correcting actions so aggressively that the gradient signal is degraded and the policy fails to improve.

A further limitation is architectural: many runtime approaches enforce a single certificate type and compose them sequentially. ATACOM~\cite{liu2021constraint,Liu2024TRO} projects actions onto the constraint manifold tangent space but lacks energy-based certificates, offering no guarantee on rotational energy dissipation. Lyapunov-based safe policy optimization~\cite{chow2019lyapunov} enforces energy dissipation but treats constraints independently, so satisfying one may violate another. The Barrier-Lyapunov Actor-Critic~\cite{zhao2023blac} integrates both but requires constrained actor updates and a backup controller, coupling safety enforcement to the learning algorithm. None jointly optimize over tilt \emph{and} Lyapunov constraints in a single step, leaving sequential composition as the only option, with the feasibility issues this entails.

This paper presents \textbf{CALOS} (Control-Affine Lyapunov
On-manifold Safety), a runtime safety layer that addresses
these limitations. CALOS unifies attitude-tilt constraints and
a Lyapunov stability condition into a single quadratic program (QP), producing the
minimum-norm torque correction at each control step. The
layer is transparent to the learning algorithm, here Proximal
Policy Optimization (PPO)~\cite{Schulman2017}, and compatible with any
policy gradient method.

\section{Problem Formulation}
\label{sec:problem}

We consider quadrotor trajectory tracking: a learning-based controller must follow a time-varying reference $p_d(t)$ while respecting attitude constraints. The quadrotor uses the Combined Thrust and Body Torques (CTBT) representation~\cite{mahony2012quadrotor}: the translational and rotational dynamics are:
\begin{align}
m\dot{v} &= mge_3 - TRe_3 + F_d, \label{eq:trans}\\
J\dot{\Omega} &= \tau - \Omega\times J\Omega, \label{eq:rot}
\end{align}
where $m$ is the mass, $v$ the inertial velocity, $R$ the rotation matrix from body to world frame, $F_d$ the aerodynamic drag, $J\in\mathbb{R}^{3\times 3}$ the inertia matrix, $\Omega\in\mathbb{R}^3$ the body angular velocity, and $\tau\in\mathbb{R}^3$ the body torque vector. The CTBT is input-affine which is essential for linear constraints on the input.

The DRL policy outputs $u_0 = (T,\,\tau_0)$, where $T$ is the collective thrust and $\tau_0\in\mathbb{R}^3$ is the nominal body torque. The safety layer receives $\tau_0$ and the current state $x = (g_b, \Omega)$, comprising the body-frame gravity vector and angular velocity, and returns a corrected torque:
\begin{equation}
\tau = \mathcal{S}(x, \tau_0),
\label{eq:safety_op}
\end{equation}
where $\mathcal{S}$ denotes the safety operator, leaving thrust unmodified to preserve vertical authority and avoid coupling with altitude regulation.

\section{Safety Layer Design}
\label{sec:safety}

\subsection{Tilt constraints and Predictive Tilt Projection}

Attitude tilt is expressed via the gravity direction in body frame, $g_b = R^\top e_3$, where the subscript $b$ denotes the body-frame representation, avoiding Euler-angle singularities. We impose roll and pitch limits $\phi_{\max} = \theta_{\max} = 60^\circ$. Using a symplectic Euler discretization of $\dot g_b = g_b\times\Omega$ and the rotational dynamics, the predicted $g_{b,t+1}$ depends affinely on $\tau$~\cite{liu2021constraint}:
\begin{equation}
g_{b,t+1} \approx g_{\text{const}}(x_t) + M(x_t)\,\tau,
\label{eq:gb_affine}
\end{equation}
where $g_{\text{const}}$ collects all $\tau$-independent terms, and $M(x_t) = \Delta t^2 [g_b]_\times J^{-1}$ is the sensitivity matrix mapping torque to changes in the body-gravity vector; $[g_b]_\times$ is the skew-symmetric matrix of $g_b$. Imposing $|g_{b,x}| \le \sin\theta_{\max}$, $|g_{b,y}| \le \sin\phi_{\max}$ yields four linear inequalities:
\begin{equation}
A_{\text{PTP}}(x)\,\tau \le b_{\text{PTP}}(x),
\label{eq:tilt_ineq}
\end{equation}
where
\begin{equation}
A_{\text{PTP}} = \begin{bmatrix} M_x \\ -M_x \\ M_y \\ -M_y \end{bmatrix}\!\in\mathbb{R}^{4\times 3},\quad
b_{\text{PTP}} = \begin{bmatrix} \sin\theta_{\max} - g_{\text{const},x} \\ \sin\theta_{\max} + g_{\text{const},x} \\ \sin\phi_{\max} - g_{\text{const},y} \\ \sin\phi_{\max} + g_{\text{const},y} \end{bmatrix}\!\in\mathbb{R}^{4},
\label{eq:ptp_explicit}
\end{equation}
with $M_x$ and $M_y$ denoting the first and second rows of $M(x_t)$. The vector $b_{\text{PTP}}$ encodes the tilt bounds offset by $g_{\text{const}}$.

The \textbf{Predictive Tilt Projection (PTP)} enforces~\eqref{eq:tilt_ineq} via an ATACOM-inspired~\cite{liu2021constraint} iterative correction. 

Let $v_+(\tau)=\max\{A_\text{PTP}\,\tau-b_\text{PTP},\,0\}$; the torque is corrected as: \begin{equation}
\tau \leftarrow \tau - A_{\text{PTP}}^{\dagger}\,k\,v_+(\tau)
\label{eq:damped_ps}
\end{equation} where $k>0$ is a tunable correction gain and $A_{\text{PTP}}^{\dagger} = A_{\text{PTP}}^\top(A_{\text{PTP}}A_{\text{PTP}}^\top+\lambda I)^{-1}$ is the damped pseudo-inverse of $A_{\text{PTP}}$, with damping factor $\lambda>0$ ensuring numerical conditioning.

\subsection{Lyapunov constraint and scaling}

We define a Lyapunov candidate \begin{equation}V(x) = \tfrac{1}{2} q_g \|e_g\|^2 + \tfrac{1}{2} q_\omega \|\Omega\|^2\end{equation} on tilt error $e_g = g_b - g_{d,b}$ and angular velocity $\Omega$, and enforce $\dot V(x,\tau) \le -cV(x)$ with decay rate $c > 0$. Since the rotational dynamics are control-affine, $\dot V$ is affine in $\tau$, yielding:
\begin{equation}
A_L(x)\,\tau \le b_L(x),
\label{eq:lyap_ineq}
\end{equation}
where $A_L = q_\omega \Omega^\top J^{-1}\in\mathbb{R}^{1\times 3}$ captures the sensitivity of rotational energy to applied torque, and $b_L = -cV(x) - B_L(x)\in\mathbb{R}$, with $B_L(x) = q_g\,e_g^\top(g_b\times\Omega) - q_\omega\,\Omega^\top J^{-1}(\Omega\times J\Omega)$ collecting the tilt-error coupling and the gyroscopic terms.

As a standalone operator, the \textbf{Lyapunov scaling}, enforces~\eqref{eq:lyap_ineq} by reducing torque magnitude: $\tau = s\,\tau_0$, $s\in[0,1]$. 

\subsection{Sequential composition}

A sequential \textbf{Cascade} composition of tilt projection, Lyapunov scaling and tilt projection check, does not guarantee joint feasibility. When the tracking error is large, $V(x)$ grows and scaling drives $s\to 0$, nullifying the torque when the controller needs maximum to recover the large error. This motivates our unified formulation.

\subsection{CALOS: unified formulation}

CALOS stacks~\eqref{eq:tilt_ineq} and~\eqref{eq:lyap_ineq} into a single system of $m{=}5$ linear constraints:
\begin{equation}
A(x) =
\begin{bmatrix}
A_{\text{PTP}}(x)\\
A_L(x)
\end{bmatrix}\!\in\mathbb{R}^{5\times 3},\quad
b(x) =
\begin{bmatrix}
b_{\text{PTP}}(x)\\
b_L(x)
\end{bmatrix}\!\in\mathbb{R}^{5}.
\label{eq:stack}
\end{equation}
The safety layer solves:
\begin{equation}
\tau^\star = \arg\min_{\tau\in\mathbb{R}^3}\; \tfrac{1}{2}\|\tau - \tau_0\|^2
\quad\text{s.t.}\quad A(x)\tau \le b(x),
\label{eq:qp}
\end{equation}
which admits a unique solution whenever the feasible set is non-empty.

\smallskip\noindent CALOS can be solved in two ways. \textbf{CALOS-P} is a projected approximation that solves~\eqref{eq:qp} approximately by applying the same damped pseudo-inverse correction~\eqref{eq:damped_ps} to the unified system $(A,b)$, enforcing all five constraints jointly without guaranteeing the exact minimum-norm solution, trading exactness for speed in large batches, which directly reduces training time when the safety layer must be evaluated across thousands of parallel environments at every timestep.

\smallskip\noindent\textbf{CALOS-QP} is an exact solver that exploits the fact that $\tau\in\mathbb{R}^3$ is three-dimensional, so at most three linearly independent constraints can be simultaneously active. This yields $\sum_{k=0}^{3}\binom{5}{k} = 26$ candidate active sets. For each candidate $I$, the Karush--Kuhn--Tucker optimality conditions give $\tau = \tau_0 - A_I^\top \lambda$ and $(A_I A_I^\top)\lambda = A_I \tau_0 - b_I$~\cite{activeset}. A candidate is feasible if the dual variables satisfy $\lambda \ge 0$ and the resulting torque satisfies all constraints $A\tau \le b$; among all feasible candidates, we select the one minimizing $\|\tau - \tau_0\|^2$.If none exists, the uncorrected policy action $\tau_0$ is applied and the actuators are clamped to their physical limits, ensuring that the system remains within hardware-safe bounds even when the QP has no feasible solution. The enumeration is fully parallelizable, essential for modern DRL training where the safety layer must scale to thousands of environments without becoming a bottleneck for the training pipeline.

\section{Experiments}
\label{sec:experiments}

\subsection{Setup}
All agents are trained with PPO~\cite{Schulman2017} using the DRL library SKRL on 4096 parallel environments in NVIDIA Isaac Lab for ${\sim}10^9$ timesteps. The policy is a two-layer MLP 64 units each, with ELU activation and Gaussian output. The observation $o_t\in\mathbb{R}^{15}$ comprises body-frame
velocity $v_b$, angular velocity $\Omega_b$, projected gravity $g_b$, desired
position $p_{d,b}$, and desired velocity $v_{d,b}$, all in $\mathbb{R}^3$. A reward curriculum progressively tightens the tracking precision from 0.50\,m to 0.15\,m throughout training, requiring increasingly accurate position following as the policy improves.

Six variants are compared: \textbf{PPO}, \textbf{PTP}, \textbf{Lyapunov scaling}, \textbf{Cascade}, \textbf{CALOS-P}, \textbf{CALOS-QP}. We evaluated the six variants over 200 episodes on three different trajectories. The first is the training Lissajous $\mathcal{L}_1$, with amplitude of 1\,m and period of 5\,s. The second is a harder Lissajous $\mathcal{L}_2$, with amplitude of 1.5\,m and 43\% higher peak velocity than the training reference. The third is a circle $\mathcal{C}$ with radius $r=1$\,m and angular rate $\omega_c=1.777$\,rad/s, chosen so that the tangential speed matches the peak velocity of the training reference.$\mathcal{L}_2$ and $\mathcal{C}$ are \emph{never seen during training} and are tested with a large initial position offset. A mean lateral error exceeding 0.20\,m is considered a tracking failure.

\subsection{Tracking Performance}
\begin{table}[!htbp]
\centering
\caption{Mean lateral error $\overline{e}_{xy}$ [m] across evaluation conditions.}
\label{tab:tracking}
\setlength{\tabcolsep}{4pt}
\begin{tabular}{lccc}
\toprule
Agent & $\mathcal{L}_1$ & $\mathcal{L}_2$ (unseen) & $\mathcal{C}$ (unseen) \\
\midrule
PPO            & 0.157 & 0.171 & 0.165 \\
PTP            & 0.081 & 0.107 & \textbf{0.040} \\
Lyapunov       & 0.127 & 0.256 & 0.569 \\
Cascade        & 0.108 & 0.243 & 0.352 \\
CALOS-P        & \textbf{0.062} & \textbf{0.076} & 0.043 \\
CALOS-QP       & 0.069 & 0.113 & 0.042 \\
\bottomrule
\end{tabular}
\end{table}

Table~\ref{tab:tracking} summarizes tracking accuracy. CALOS-P and CALOS-QP reduce lateral error by 55--60\% on $\mathcal{L}_1$ relative to PPO. On the unseen trajectories with large initial offset, Lyapunov and Cascade fail to track the reference due to the torque nullification discussed in Section~\ref{sec:safety}, while both CALOS variants maintain errors below 0.08\,m.

\begin{figure}[!htbp]
\centering
\includegraphics[width=0.95\columnwidth]{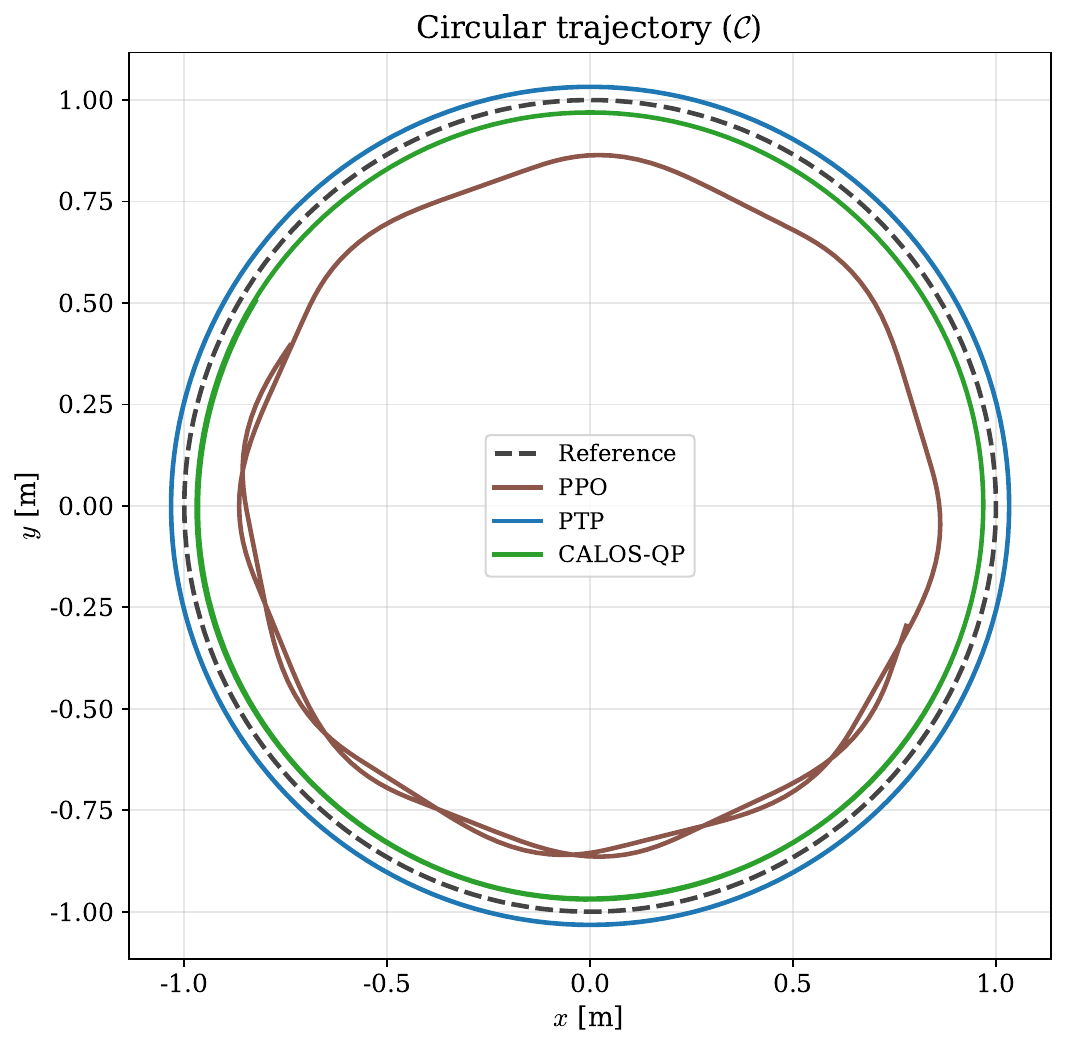}\\[4pt]
\includegraphics[width=0.95\columnwidth]{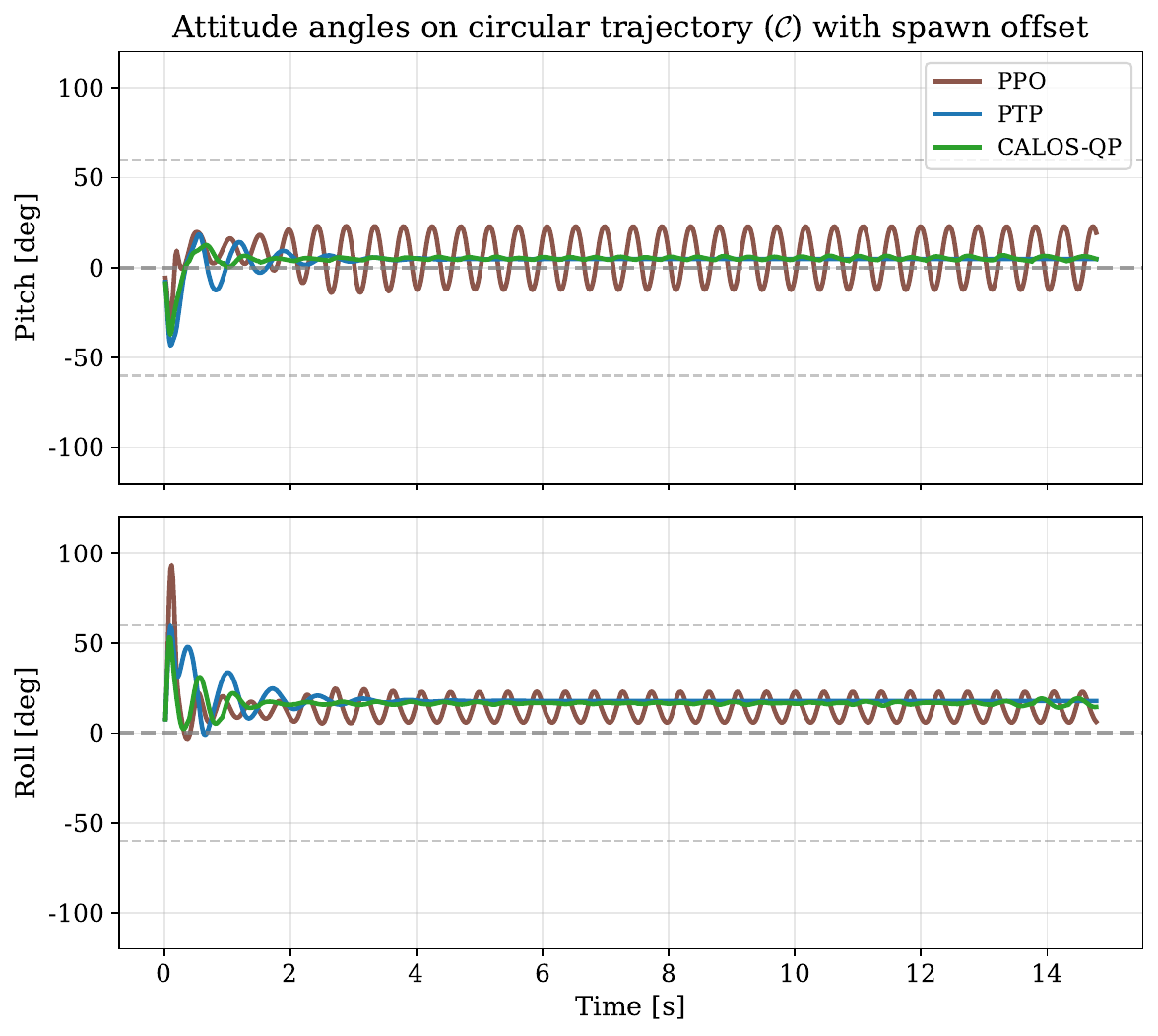}
\caption{Circular trajectory $\mathcal{C}$ with large initial offset. Top: $xy$-plane tracking (reference dashed). Bottom: roll and pitch; dashed lines mark the $\pm60^\circ$ constraint. PTP is included to show the effect of the Lyapunov constraint: CALOS-QP settles faster due to active energy dissipation.}
\label{fig:circle}
\end{figure}

Figure ~\ref{fig:circle} shows results on the circular trajectory $\mathcal{C}$ . The
constant curvature of this reference produces a steady-state
attitude regime once the initial transient has settled, making
it straightforward to distinguish the transient recovery phase
from nominal tracking and to identify differences in constraint
enforcement across methods. The trajectory plot (top)
confirms that CALOS-QP tracks the reference closely while
PPO exhibits visible drift. In the attitude plot (bottom), PTP
is also shown to highlight the effect of the Lyapunov component:
CALOS-QP extends PTP by adding the energy dissipation
constraint ~\eqref{eq:lyap_ineq} to the unified QP, and this additional
term visibly accelerates the convergence of roll and pitch toward
the steady-state regime. PTP, which enforces only tilt
bounds, exhibits larger oscillations and a slower settling, because
it has no mechanism to actively dissipate rotational energy
during the recovery phase. CALOS-QP, by jointly enforcing
both tilt and Lyapunov constraints, damps the angular
transient more aggressively while still respecting the attitude
bounds.

\subsection{Safety Analysis}

Table~\ref{tab:safety} reports attitude safety metrics on $\mathcal{L}_1$ evaluated with a large initial position offset, a condition never encountered during training, designed to stress-test the safety layer under extreme tracking error. The roll and pitch constraint enforced by CALOS is set at $60^\circ$.

\begin{table}[!htbp]
\centering
\caption{Roll-violation statistics on $\mathcal{L}_1$ with large initial offset (200 episodes, 60$^\circ$ limit). Avg and Max refer to consecutive steps in violation.}
\label{tab:safety}
\setlength{\tabcolsep}{3.5pt}
\begin{tabular}{lcccc}
\toprule
Agent & $\max|\phi|$ & Viol.\ eps. & Avg steps & Max steps\\
\midrule
PPO       & 116.9$^\circ$ & 4096 & 5.0 & 6 \\
PTP       & 68.7$^\circ$  & 1122 & 1.8 & 3 \\
Lyapunov  & 180.0$^\circ$ & 2988 & 5.2 & 27 \\
Cascade   & 175.4$^\circ$ & 1599 & 3.2 & 14 \\
CALOS-P   & 71.5$^\circ$  & 4096 & 2.6 & 3 \\
CALOS-QP  & 99.9$^\circ$  & 3069 & 4.3 & 9 \\
\bottomrule
\end{tabular}
\end{table}

CALOS-P violations last at most 3 consecutive steps, occurring only during the first simulation steps when the controller is correcting the large initial error; once the attitude enters the feasible region, no further violations occur.

In contrast, Lyapunov violations reach a maximum tilt of $180^\circ$, lasting up to 27 consecutive steps. On $\mathcal{L}_1$ \emph{without} an initial offset, CALOS-QP achieves zero roll-violation episodes, demonstrating that the safety layer fully enforces the attitude constraints under nominal conditions. The residual violations observed under large initial offsets could be further reduced by including such high-error conditions in the training distribution.

\subsection{Internalized Safe Behavior}

The policy trained with CALOS retains safer and more accurate behavior even without the safety layer at test time, as shown in Table~\ref{tab:layersoff}: the mean lateral error is 55--74\% lower than PPO. The constraint
projection prevents unrecoverable attitudes during training,
so all samples come from states where meaningful learning
occurs. This restricts exploration to safe regions, accelerating
convergence and improving data efficiency. Moreover, CALOS formulation preserves sufficient freedom within the
feasible set, avoiding suboptimal policies: the policy internalizes
the constraint boundaries as part of its nominal behavior while matching or exceeding the tracking performance of the unconstrained baseline.

\begin{table}[!htbp]
\centering
\caption{Tracking with the safety layer disabled at test time.}
\label{tab:layersoff}
\setlength{\tabcolsep}{4pt}
\begin{tabular}{llcc}
\toprule
Traj. & Agent & $\overline{e}_{xy}$ [m] & $\overline{R}$ \\
\midrule
\multirow{3}{*}{$\mathcal{L}_1$} & PPO        & 0.157 & 93.2 \\
                                   & CALOS-P (layer off) & 0.062 & 100.9 \\
                                   & CALOS-QP (layer off) & 0.067 & 100.9 \\
\midrule
\multirow{3}{*}{$\mathcal{C}$}    & PPO        & 0.165 & 90.1 \\
                                   & CALOS-P (layer off) & 0.043 & 99.0 \\
                                   & CALOS-QP (layer off) & 0.041 & 100.3 \\
\bottomrule
\end{tabular}
\end{table}

\subsection{Why Sequential Methods Fail}

The failure of Lyapunov scaling and Cascade under large tracking errors is due to the multiplicative structure: $\tau = s\tau_0$ with $s\to 0$ when $V(x)$ is large, nullifying the torque precisely when maximum authority is needed. CALOS encodes the Lyapunov condition as $A_L(x)\tau \le b_L(x)$, defining in $\mathbb{R}^3$ the set of all torque vectors whose component along $A_L$ does not exceed the dissipation budget $b_L$. The QP can project the torque to satisfy dissipation while preserving a large component along the recovery direction.

Across all experimental conditions---all agents, trajectories, with and without initial offset---the CALOS-QP solver \emph{always} found a feasible solution to~\eqref{eq:qp}. The fallback to $\tau_0$, i.e., applying the uncorrected policy action, was never triggered, confirming that the joint tilt-Lyapunov constraint set remained non-empty throughout all tests.

\section{Conclusions}

We presented CALOS, a runtime safety layer that unifies tilt and Lyapunov constraints into a single low-dimensional QP for quadrotor attitude control under DRL.The method enforces constraints in real time with minimal intervention—achieving zero attitude violations on the training trajectory and at most 3 consecutive violation steps under large perturbations—while reducing lateral tracking error by 55–74\% relative to unconstrained PPO. The QP feasible set was never empty across all tests, meaning the fallback to the uncorrected policy action was never triggered. Moreover, policies trained with CALOS internalize safe behavior, retaining comparable tracking improvement even with the safety layer disabled, confirming that the constrained exploration improves convergence and data efficiency without sacrificing optimality.

Future work includes: (i) learned, task-aware Lyapunov certificates~\cite{dai2021lyapunov,Wu2023NeurIPS}: the current quadratic certificate drives toward the zero-tilt equilibrium, causing unnecessary interventions when the drone tracks at a non-zero but feasible attitude; a learned certificate that decreases toward the current reference would eliminate this; (ii)~extensions to non-control-affine dynamics, since higher-fidelity models with rotor-speed dynamics or blade-flapping effects break the affine assumption that CALOS requires; (iii) sim-to-real transfer with domain randomization and data-driven safety filters~\cite{Wabersich2023} to handle model uncertainty on physical hardware.

{\small
\bibliographystyle{ieeetr}
\bibliography{paper}
}

\end{document}